\documentclass[a4paper,12pt]{IEEEtran}

\usepackage{cite}
\usepackage{amsmath,graphicx}
\usepackage{multirow}
\usepackage{epsfig, subfigure}
\usepackage[update,prepend]{epstopdf}
\usepackage{caption}
\usepackage{color}
\usepackage{amsfonts}
\usepackage{algorithmic}
\usepackage{algorithm}

\newtheorem{property}{\sc \bf Property}[section]
\newtheorem{propos}{\sc \bf Proposition}[section]
\newtheorem{theor}{\sc \bf Theorem}[section]
\newtheorem{corr}{\sc \bf Corollary}[section]
\newtheorem{remark}{\sc \bf Remark}[section]
\newtheorem{definition}{\sc \bf Definition}[section]
\newtheorem{example}{\sc \bf Example}[section]

\newcommand{\etal}{et al.\!}
\newcommand{\eg}{e.g.\!}
\newcommand{\ie}{i.e.\!}

\ifCLASSINFOpdf
\else
\fi
\begin{document}

\title{Zero-Shot Object Recognition System \\ based on Topic Model}

\author{Wai~Lam~Hoo
       and~Chee~Seng~Chan
\thanks{The authors are with the Centre of Image and Signal Processing, Faculty of Computer Science and Information Technology, University of Malaya, 50603 Kuala Lumpur, MALAYSIA.} \thanks{Corresponding author: C.S. Chan (email: cs.chan@um.edu.my).}}


\markboth{To Appear in IEEE THMS}%
{Shell \MakeLowercase{\textit{et al.}}: Bare Demo of IEEEtran.cls for Journals}



\IEEEtitleabstractindextext{%
\begin{abstract}
Object recognition systems usually require fully complete manually labeled training data to train the classifier. In this paper, we study the problem of object recognition where the training samples are missing during the classifier learning stage, a task also known as zero-shot learning. We propose a novel zero-shot learning strategy that utilizes the topic model and hierarchical class concept. Our proposed method advanced where cumbersome human annotation stage (\ie~attribute-based classification) is eliminated. We achieve comparable performance with state-of-the-art algorithms in four public datasets: PubFig ($67.09\%$), Cifar-100 ($54.85\%$), Caltech-256 ($52.14\%$), and Animals with Attributes ($49.65\%$) when unseen classes exist in the classification task.
\end{abstract}

}

\maketitle

\IEEEdisplaynontitleabstractindextext

%
\IEEEpeerreviewmaketitle

\section{Introduction}

Object classification from natural images is useful in content-based image retrieval, video surveillance, robot localization and image understanding. According to Lampert \etal~\cite{Lampert2009}, humans are able to distinguish between at least 30,000 relevant classes. However, training conventional object detectors for all these classes would require millions of well-labeled training images and is likely out of reach for years to come.

As such, the zero-shot learning paradigm \cite{Lampert2009, Parikh2011, Kumar2009, rohrbach2011evaluating,frome2013devise, mensink2012metric, Hoo2013} is motivated from the human ability to learn and abstract from examples, and the capability to describe completely unseen classes (\ie~ training classes are not available during training of the object detector) from existing (known) classes. For instance, \cite{Lampert2009, Parikh2011, Kumar2009, rohrbach2011evaluating} recognize a set of unseen objects using a list of high-level attributes that serve as an intermediate layer in the classifier cascade. The attributes enable those systems to recognize the object classes, even without a single training example. Others like \cite{frome2013devise, mensink2012metric} use semantic relationships from different reference classes to predict the unseen classes. Though promising results were obtained, all these aforementioned approaches require either extensive human supervision to build the attributes, or a tight semantic relationship between the unseen classes and the training classes. 

\begin{figure*}[t]
\centering
\hfill
\subfigure[Weakly Supervised Learning \cite{fei2005bayesian, Griffin2007,Bosch2007,Moosmann_Nowak_Jurie_2008}]{\includegraphics[height=0.18\linewidth, width=0.28\linewidth]{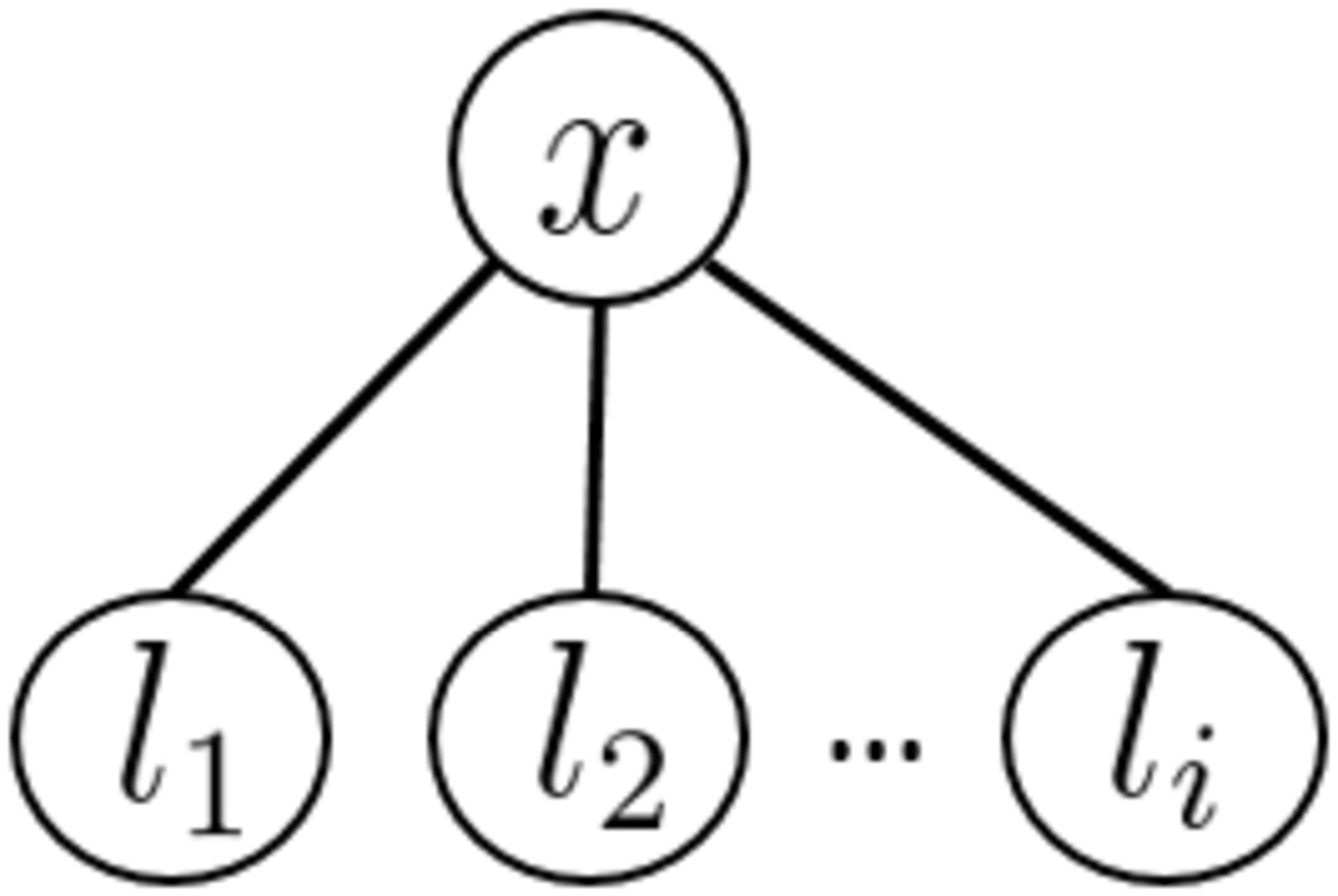}}
\hfill
\subfigure[Attributes \cite{Parikh2011,Kumar2009,ferrari2007learning,Lampert2009,Lampert2014, fu2012attribute, Fu2014Learning} or tags \cite{guillaumin2010multimodal}]{\includegraphics[height=0.25\linewidth, width=0.25\linewidth]{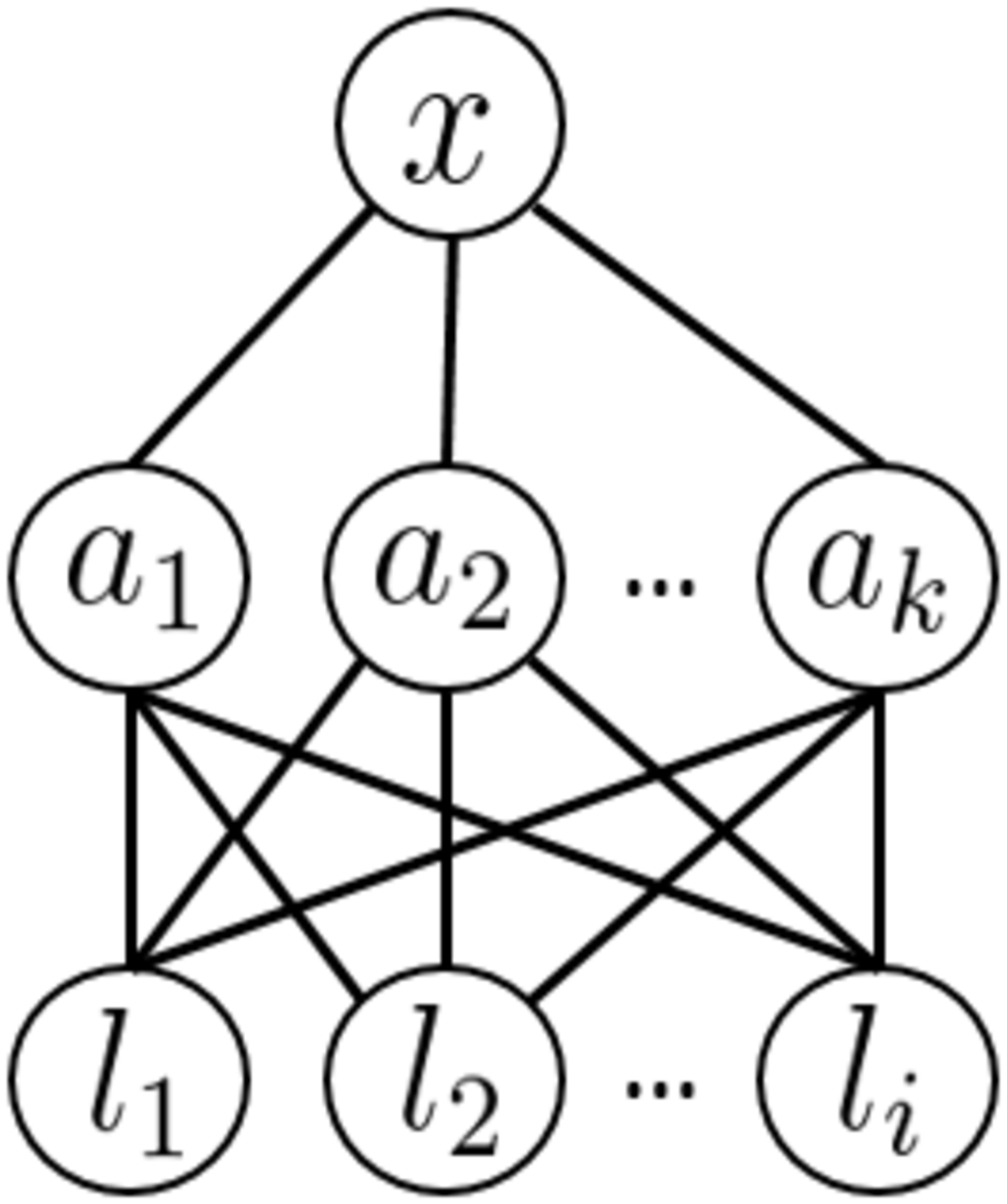}}
\hfill
\subfigure[Hierarchical class (HiC) concept]{\includegraphics[height=0.25\linewidth, width=0.25\linewidth]{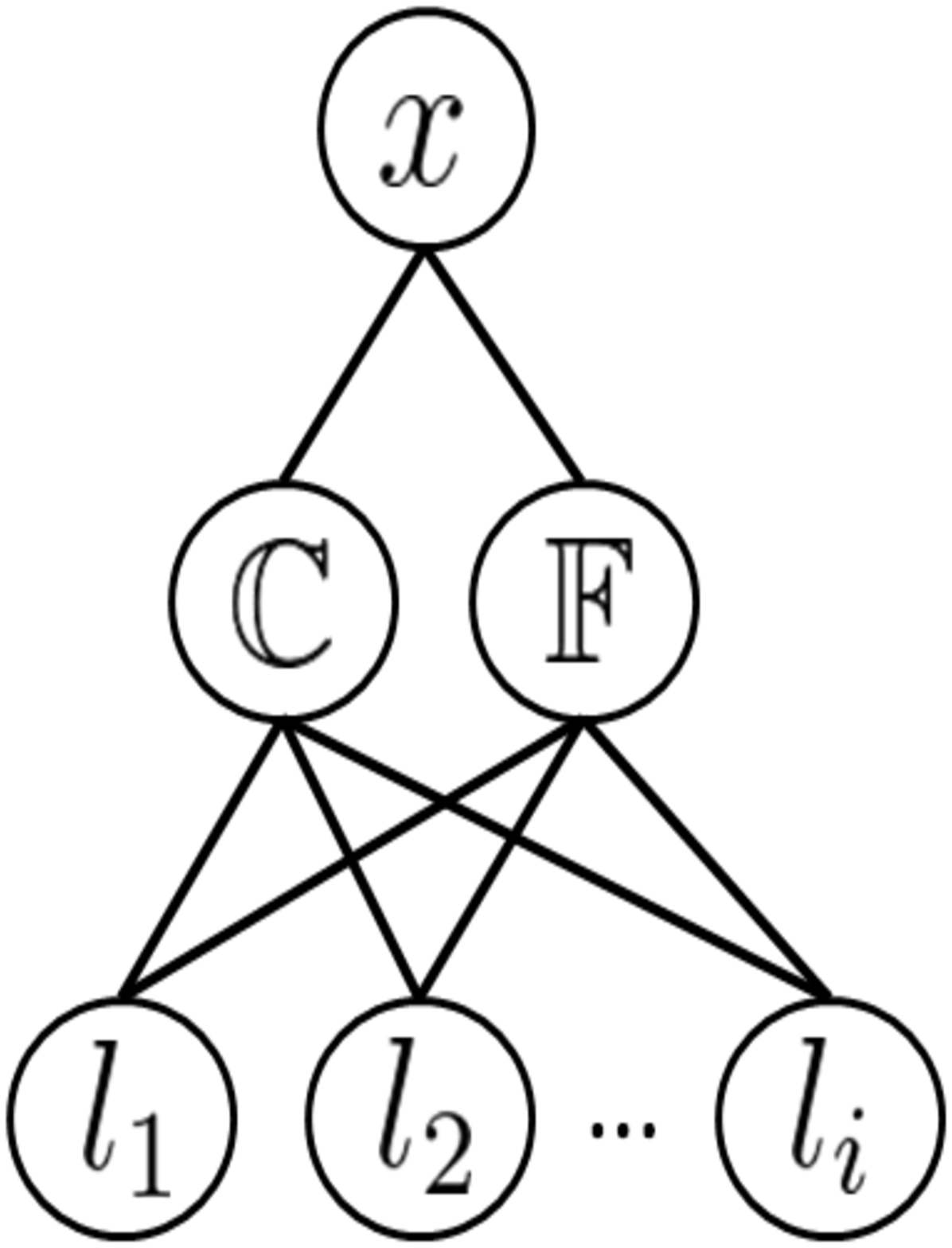}}
\caption{Comparison between (a) Weakly supervised learning as conventional learning algorithms; (b) Learning classifier by including association between images with attributes, or tags, respectively; (c) Hierarchical Class (HiC) concept.}
\label{fig:intro}
\end{figure*}

In this paper, we propose 1) {\it topic model} to replace the attributes \cite{Lampert2009, Parikh2011, Kumar2009, rohrbach2011evaluating} so that extensive human supervision is no longer required, and 2) the {\it Hierarchical Class (HiC) concept} to relate the unseen classes to the existing seen classes. The HiC concept has a loose relationship image hierarchy compared to \cite{frome2013devise, mensink2012metric}. Our framework starts with building a Bag-of-Words (BoW) model using the image features from the (small amount of) seen (available) classes. Herein, the HiC concept is utilized to build the codebook. A topic model (here we employ the probabilistic Latent Semantic Analysis (pLSA)) is learned using the generated BoW model. Based on the learned pLSA model and HiC concept, signature topics for both the seen and unseen classes are deduced (\ie~we cluster similar object classes that share visual similarity). Finally, object classification is performed using the deduced signature topics representation. Experimental results using four publicly available datasets, namely the PubFig, Cifar-100, Caltech-256 and AwA datasets have shown the effectiveness of the proposed method.  

The rest of the paper is structured as follows. Section \ref{sec:related_work} presents the related work. Section \ref{sec:method} details the proposed methodology. Section \ref{sec:exp} shows the experimental results and Section \ref{sec:discussion} presents the discussion.
  

\section{Related work}
\label{sec:related_work}

Palatucci \etal~\cite{Palatucci2009} showed that the attribute description of an instance or category is useful as a semantically meaningful intermediate representation to bridge the gap between low level features and high-level classes. Thus, the attributes facilitate transfer and zero-shot learning to alleviate issues of the lack of labeled training data, by expressing classes in terms of well-known attributes. This is followed by Lampert \etal~\cite{Lampert2009, Lampert2014} that extended the work to animal categorization by introducing Direct Attributes Prediction (DAP) and Indirect Attributes Prediction (IAP).

Unlike \cite{Palatucci2009,Lampert2009,Lampert2014}, Parikh and Grauman \cite{Parikh2011} introduced relative attributes to perform zero-shot learning. This approach captures the relationships between images and objects in terms of human-nameable visual properties. For example, the models capture that animal $A$ is `taller' than animal $B$, or subject $X$ is `happier' than subject $B$. This allows a richer language of supervision and description than the commonly used categorical (binary) attributes. Though relative attributes seem efficient for zero-shot learning, the dataset needs to be intra-class (\ie~the images in the dataset must belong to a set of object classes that are visually similar). Also, a binary or relative relationship between all classes needs to be defined beforehand. Such a process will require extensive human supervision efforts and the decision is always subjective. 

In our proposed strategy, we replace the attributes with a topic model in order to reduce the human supervision needed. Others who use topic models in zero-shot learning are \cite{Fu2014Learning, fu2012attribute}. They propose a hybrid attribute-topic model to deal with group social activities. Specifically, they define three unique attributes: user-defined, latent class-conditional, and latent generalized free attributes. These attributes are learned jointly in a semi-latent attribute space, and as the multi-modal latent attribute topic model (M2LATM). The motivation is to reduce the annotation effort through the introduction of the latent attributes in their proposed framework. In contrast, our focus in this paper is on object recognition that learns the topic model directly from the BoW representations, and infers the unseen classes using the proposed HiC concept. We eliminate the time consuming human annotation process by replacing the attributes with topic models. Instead of learning the topic models on top of user-defined and latent attributes \cite{Fu2014Learning, fu2012attribute}, we choose the pLSA as our topic model because it does not require prior comparison to the Latent Dirichlet Allocation (LDA) model. We further extend the topic model representation as a mapping algorithm to object classes, so that zero-shot learning would be possible. 
 
Figure \ref{fig:intro} shows conventional solutions that associate each image with a class label \cite{fei2005bayesian, Griffin2007,Bosch2007,Moosmann_Nowak_Jurie_2008}, or further describe the image content with the association of attributes \cite{Parikh2011,Kumar2009,ferrari2007learning,Lampert2009,Lampert2014, fu2012attribute, Fu2014Learning} or image tags \cite{guillaumin2010multimodal}. These are insufficient in zero-shot learning because these attributes and tags can be redundant and not useful when too many of them are introduced. Yet, there is no specific evaluation method on "what is an effective attribute or tag". Therefore, we introduce a new codebook learning method, \ie~ the HiC concept that utilizes the hierarchical class characteristics during the codebook learning stage. This concept is inspired by \cite{silberer2013models, frome2013devise, mensink2012metric} where a set of common objects are clustered into different classes in order to deduce the relationship among them. Specifically, we integrate two different levels of image class labels, namely the Coarse Class, $\mathbb{C}$ and Fine Class, $\mathbb{F}$.
Then, this class hierarchy is learned in the topic model to identify the significant differences among the classes and improve the model prediction capability. Such an approach is better than attributes-based classification \cite{Lampert2009,Lampert2014} which are commonly applicable in inter-class problems only. The HiC concept manages to deal with both inter-class, as well as intra-class problems.

Similar work that employed the hierarchical class strategy in zero-shot learning paradigm includes Rohrbach \etal~ \cite{rohrbach2011evaluating} and Frome \etal~\cite{frome2013devise}. In both approaches, a set of frameworks on how to incorporate the semantic information from a language model/set to assist in the zero-shot learning is studied. \cite{rohrbach2011evaluating} employed WordNet and Wikipedia as the language model, and learned a similarity measure to represent the hierarchy/attributes/objectness measure between the object classes. \cite{frome2013devise} extended the idea to learn the class relationship directly from the unannotated data (\ie~visual-semantic relationship between object classes from millions of documents in Wikipedia) using the Deep Visual-Semantic Embedding Model (DeViSE). 
 
In another approach, Mensink \etal~\cite{mensink2012metric} used a different concept where a distance metric from a set of seen classes (\eg~800 seen classes) and errors for both seen and unseen classes (\eg~800 seen classes and 200 unseen classes, result in 1000-way classification) are learned. In order to classify the object classes, a Nearest Class Mean (NCM) classifier is employed. This approach does not require the semantic relationship, and manages to generalize the unseen classes in near to zero computational cost. For our proposed framework, although it is similar to the hierarchy-based knowledge transfer in \cite{rohrbach2011evaluating}, we do not need a language model to build the hierarchy. Instead, the HiC concept relates the unseen classes to the seen classes. Also, we use learnt topic model to perform the zero-shot learning, which is different from the attributes-based or direct similarity-based knowledge transfer in \cite{rohrbach2011evaluating} that uses attributes or objectness measure, and \cite{mensink2012metric} that uses metric learning.
 
\section{Approach}
\label{sec:method}

In this section, we first discuss the prerequisites of the proposed framework: BoW model and topic model. Secondly, we explain the HiC concept and detail how to perform zero-shot learning in pLSA with the HiC concept. Finally, we show the inference method for image classification purposes. 

\subsection{Codebook Representation}
\label{sec:BoW} 

To build the BoW model, we engaged the Random Forest (RF) algorithm \cite{Moosmann_Nowak_Jurie_2008,WLHoo} where a random decision tree is constructed using a random subset of the training data with replacement.  The {\bf labeled} training images at a particular node $I_{\text{node}} = \{x_i, l_i\}$ are recursively split into left node $I_{\text{left}}$ and right node $I_\text{right}$ subsets, according to a threshold $t \in T$ and a split function $f$ (Eq. \ref{Eq1}).

\begin{equation}
I_{\text{left}} = \{x_i \in I_{\text{node}} | f(x_i) < t\}, I_{\text{right}} = I_{\text{node}} \setminus I_{\text{left}},
\label{Eq1}
\end{equation}

\noindent where $x_i$ are the feature vectors from the training images and $l_i$  are the associated class labels. At each split node, random subsets of features are generated and compare to $T$. In this process, $t \in T$ that maximizes the expected information gain $\triangle E$ is selected:

\begin{equation}
\label{Information_gain}
\triangle E =E(I_{\text{node}}) -\sum_{p=\text{left},\text{right}} \frac{\mid I_p \mid}{ \mid I_{\text{node}} \mid}E(I_p) ,
\end{equation}

\noindent where $E(I) = p(l_i) \, \text{log} \, p(l_i)$, and $E(I)$ is the Shannon entropy of the probability class histogram $p(l_i)$. As such, the leafnodes of all trees in the RF form a codebook. Then, the codebook are used to quantize $I$ into BoW representation, by passing $x_i$ to each tree and count the occurrence of each leafnode.

\subsection{Topic Model}

Our model is based on a latent topic model, in particular, the pLSA model. We briefly introduce it using the terminology in our context. Suppose we are given a collection of images $D = \{d_1,\cdots,d_N\}$. 
Each image $d$ is represented by a collection of features $W = \{w_1,\cdots,w_V\}$, where it shows how frequent a particular $w_v$ is used in $d$. A word is the basic item from a codebook indexed by $\{1,2,\cdots,V\}$. A joint probability model $p(w,d)$ over $V \times N$ can be defined as: 

\begin{equation}
p(w,d) = \sum p(z)p(w|z) p(z|d),
\label{eq1}
\end{equation}

\noindent where $z \in Z = \{z_1,\cdots,z_K\}$  is a latent variable. We can further derive the document-specific word distribution $p(w|d)$ as:  

\begin{equation}
p(w|d) = \sum p(w|z) p(z|d).
\label{eq2}
\end{equation}

However, at the current setting, Eq. \ref{eq1}-\ref{eq2} could not infer the unseen classes as the algorithm needs prior knowledge about which $z$ belongs to which $c$ \cite{Sivic_Russell_Efros_Zisserman_Freeman_2005}, or a set of labeled training image $\{x_i,l_i\}$ in learning the model. In the zero-shot paradigm, such information is simply not available. In order to handle this issue, we proposed the HiC concept (discussed next), so that we can infer the unseen classes to perform zero-shot learning using the pLSA model.

\subsection{Hierarchical Class (HiC) Concept}
 
We introduced the HiC concept - a nested class concept as illustrated in Figure \ref{fig:intro}c where one image consists of two class labels (semantically related), $HiC = \{\mathbb{C}, \mathbb{F}\}$.  One has a broader visual concept, namely the Coarse Class; while the other class labels have a narrow visual concept, namely the Fine Class. Table \ref{tab:method1} shows some examples of the HiC concept. 

\begin{definition}
\label{def:Def1b}
\normalfont Coarse Class, $\mathbb{C}$ is a large concept class (parent) that shares a conceptual similarity, either physical or biological, within its own
Fine Class;
\end{definition} 

\begin{definition}
\label{def:Def1a}
\normalfont Fine Class, $\mathbb{F}$ is a specific object class and is a subset to one of the Coarse Class (child).
\end{definition}

\begin{table}[!t]
\caption{Examples of the $\mathbb{C}$ and $\mathbb{F}$ relationship in HiC concept}
\label{tab:method1}
\centering
\renewcommand{\arraystretch}{1.4}
\resizebox{0.95\linewidth}{!}{
\begin{tabular}{|l||c||c||c|}
\hline
$\mathbb{C}$ & {\bf Electrical Devices} & {\bf Building} & {\bf Water Spot} \\
\hline
\hline
& - Television & - House & - Coast \\
$\mathbb{F}$ & - Refrigerator & - Apartment & - Beach \\
& - Washing Machine & - Tall Building & - Underwater \\
\hline
\end{tabular}}
\end{table}

\begin{figure*}[!ht]
\centering
\includegraphics[height=0.45\linewidth, width=0.95\linewidth]{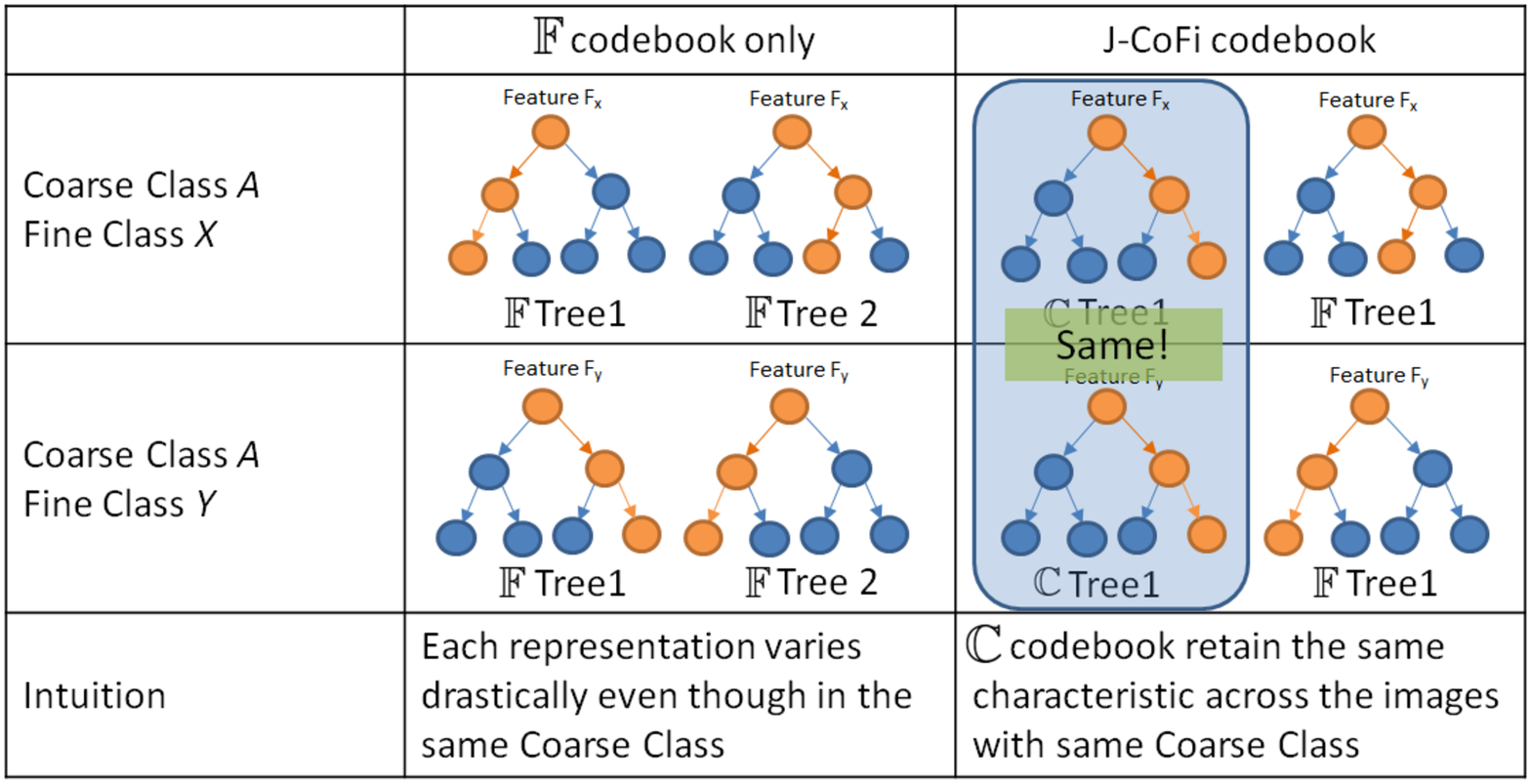}
\caption{Effects of Coarse Class ($\mathbb{C}$) on J-CoFi codebook compare with ordinary $\mathbb{F}$-based codebook. Two examples include building a random forest with 2 trees. One example has $\mathbb{F}$ trees only. Another example has 1 $\mathbb{C}$ tree and 1 $\mathbb{F}$ tree. After the trees are built, we fit feature $F_x$ and $F_y$ that have the similar $\mathbb{C}$, but different $\mathbb{F}$ into the trees. We notice significant differences in the leafnode (orange nodes indicate the path choice of the feature $F$) for a $\mathbb{F}$-based codebook. However, with the help of $\mathbb{C}$ in the J-CoFi codebook, the feature $F$ with the same $\mathbb{C}$ will have similar path choice in the $\mathbb{C}$ Tree, but a different one in the $\mathbb{F}$ tree. With this, we can build a BoW model that retains similarities for the images that belong to the same $\mathbb{C}$. }
\label{fig:method3}
\end{figure*} 

\subsubsection{Codebook Representation in HiC concept}
\label{subsec:codebook extension}

Using the HiC concept, we have three new codebook representations, that are 1) Coarse ($\mathbb{C}$) or Fine ($\mathbb{F}$), 2) Joint Coarse-Fine (J-CoFi) and 3) CoarseFine (CoFi). We next explain their properties.

\begin{property}
\label{def:Def1b1}
(Coarse ($\mathbb{C}$) or Fine ($\mathbb{F}$)). The $\mathbb{C}$ and $\mathbb{F}$ codebooks are similar to the initial RF learning described in Section \ref{sec:BoW}, except that we substitute $l_i$ in the Shannon entropy with $\mathbb{C}_i$ or $\mathbb{F}_i$, respectively. We illustrate in Figure \ref{fig:method3} that utilizing only the $\mathbb{F}$ codebook is not an optimum setting as each of the codebook representations varies drastically although they belong to the same $\mathbb{C}$. Therefore, we built a variant, namely the J-CoFi.
\end{property} 

\begin{property}
\label{def:Def1b2}
(Joint Coarse-Fine (J-CoFi)). The J-CoFi codebook strategy adapts both $\mathbb{C}$ and $\mathbb{F}$ information during the RF learning. Specifically, we denote the total number of trees as $R$. If one uses $r$ of $\mathbb{C}$ trees that govern the similarity between $x_i$ with the same $\mathbb{C}_i$, and $R-r$ of $\mathbb{F}$ trees that distinguish those $x_i$ within its associated $\mathbb{C}$, this will result in a BoW model that has a similar histogram shape for codebook bins that are created by $\mathbb{C}$ trees. Hence, it eliminates the limitations in {\bf Property \ref{def:Def1b1}}.
\end{property} 

\begin{table}[!t]
\caption{Comparison between (Coarse ($\mathbb{C}$) or Fine ($\mathbb{F}$)) vs. (Joint Coarse-Fine (J-CoFi)) codebook learning strategy}
\label{tab:method4}
\centering
\renewcommand{\arraystretch}{1.9}
\resizebox{0.95\linewidth}{!}{
\begin{tabular}{|l||c|c||c||c|}
\hline
{\bf Codebook Type} & {$\mathbb{F}$} & {$\mathbb{C}$} & {J-CoFi} & Shannon Entropy ($E(I)$)\\
\hline
\hline
$\mathbb{C}$ Tree 	& No 	& Yes 	& Yes 	& $p(\mathbb{C}) \, \text{log} \, p(\mathbb{C})$\\
\hline
\hline
$\mathbb{F}$ Tree 	& Yes 	& No 	& Yes 	& $p(\mathbb{F}) \, \text{log} \, p(\mathbb{F})$\\
\hline
\end{tabular}}
\end{table}

Table \ref{tab:method4} summarizes the difference between {\bf Property \ref{def:Def1b1} - \ref{def:Def1b2}}. There still exist limitations in the {\bf Property \ref{def:Def1b1} - \ref{def:Def1b2}} when Eq. \ref{Information_gain} is employed to compute $\triangle E$. That is, at one time, one could only optimize either $\mathbb{C}_i$ or $\mathbb{F}_i$ during the RF tree node splitting, and so we introduce the CoFi codebook ({\bf Property \ref{def:Def1b3}}) to handle this limitation.

\begin{property}
\label{def:Def1b3}
(CoarseFine (CoFi)).  The CoFi is proposed to learn the trees in such a way that utilizes both the $\mathbb{C}_i$ and $\mathbb{F}_i$, simultaneously in the RF tree node splitting. Specifically, we modified Eq. \ref{Information_gain} so for each CoFi tree, we consider the total maximum $\triangle E$ from $\mathbb{C}_i$ and $\mathbb{F}_i$ simultaneously for each split node as $\triangle E_\text{total}$:

\begin{equation}
\triangle E_\text{total} = \sum_{\mathbb{F},\mathbb{C} \in c } [E(I_{c}) - \sum_{p=\text{left},\text{right}} \frac{\mid I_{c_p} \mid}{ \mid I_n \mid}E(I_p)].
\end{equation}

\noindent and the splits that maximize the $\triangle E_\text{total}$ will be selected. 
\end{property}

\subsection{Zero-shot learning in pLSA with HiC concept}
\label{subsec:zero-shot}

In order to perform the zero-shot learning using the HiC concept, we denote a seen class as $s \in S$ and an unseen class as $u \in U$, where $\{S,U\} \subset C$. As such, we collect a set of seen classes pair $\alpha_u$ for each $u$ that associate $u$ to a pair of seen classes $s$ which belongs to the same $\mathbb{C}$:

{\small
\begin{equation}
\alpha_u = \{(g,h) \in \mathbb{F}, \mathbb{F} \subset \mathbb{C} | g \sim u \sim h\}
\label{eq:alpha_u}
\end{equation}}

\noindent where $ \{g,h\} \in S$ and $\sim$ indicates conceptual similarity between $c$ (\ie~as described in Definition \ref{def:Def1b} and in \cite{Parikh2011}). In the pLSA model, we introduce a novel mapping algorithm namely topic sets, $\mathbb{T}$ that indicate index of $z$. Each $c_m$ will associate with specific $\mathbb{T}_m$, which creates a relationship between $z$ and $c_m$. Our idea is that the unseen class $u$ that could be related to a pair of unseen classes $s$ (\ie~in this case are $g$ and $h$) will have high similarity for their respective $\mathbb{T}$. Therefore, we could relate $u$ by defining $\mathbb{T}_u$ that satisfies the conditions of $\mathbb{T}_g \sim \mathbb{T}_u \sim \mathbb{T}_h$ and $(g,h) \in S$. We denote $\mathbb{T}_s$ as the signature topic set for the seen class $s$ as:

\begin{equation}
\mathbb{T}_s = \operatorname*{arg \, max}_{\mathbb{T}_m} \; \sum_{k \in \mathbb{T}_m} p(z_k|d_m),
\label{eq:T}
\end{equation}

\noindent where the size of $M$ is $2^K$, and $p(z_k|d_m)$ is a class-specific topic distribution that is used to determine $\mathbb{T}_m$ for every $c_m$:

\begin{equation}
p(z_k|d_m) = \frac{\sum_{n \subset m} p(z_k|d_n)}{\sum_{m} p(z_k|d_m)}
\end{equation}

\noindent where $\mathbb{T}_u$ is inferred as the union of the $\mathbb{T}_s$ pairs ($\mathbb{T}_g$ and $\mathbb{T}_h$) to achieve zero-shot learning.  
Taking $K = 3$ as an example, the size of $M$ is $8$ ([0 0 1], [0 1 0], [1 0 0], [0 1 1], [1 0 0], [1 0 1], [1 1 0], [1 1 1]), where $1$ indicates the signature topic(s) and vice versa. Ideally, if $\mathbb{T}_{g}$ is [0 0 1] and $\mathbb{T}_{h}$ is [1 0 0], then $\mathbb{T}_u$ is [1 0 1]. 

Finally, given a test class $c'_m$, it can be predicted by evaluating: 

\begin{equation}
p(c'_m|d_\text{test}) = \frac{ \sum_{k \in \mathbb{T}_{c'_m}} p(z_k|d_{\text{test}})}{\sum_{m} p(c_m|d_{\text{test}}) }.
\label{final}
\end{equation}

Algorithm \ref{Algo:Modeling} summarizes the proposed framework.

\begin{algorithm}[t]
\caption{Proposed Framework}
\begin{algorithmic} 

\REQUIRE A set of labeled training images $\{x_i,l_i\}$, HiC concept, identify unseen classes $U$ and seen classes $S$.
\ENSURE All parameters are set: number of trees $R$, number of leafnodes per tree, number of topics $K$ and number of unseen class $q$.
\STATE 1. Learn RF codebook using $\{x_i,l_i\}$, either using $\mathbb{F}$ codebook, J-CoFi codebook or CoFi codebook ({\bf Property \ref{def:Def1b1} - \ref{def:Def1b3}}).
\STATE 2. Build BoW histogram based on the codebook in Step 1.
\STATE 3. Learn pLSA model using the BoW histogram.
\STATE 4. Find $\alpha_u$ for each $U$ based on Eq. \ref{eq:alpha_u}.
\STATE 5. Calculate the signature topic sets $\mathbb{T}_s$ for each $S$ as to Eq. \ref{eq:T}.
\STATE 6. Randomly pick $\alpha_u$ to relate $u$ with $s$ in terms of $\mathbb{T}_g \sim \mathbb{T}_u \sim \mathbb{T}_h, (g,h) \in S$.
\STATE 7. Calculate the signature topic sets $\mathbb{T}_u$ for each $U$ as to Eq. \ref{eq:T}.
\STATE 8. Classification for test class $c'_m$ using Eq. \ref{final}.
\end{algorithmic}
\label{Algo:Modeling}
\end{algorithm}

\section{Results}
\label{sec:exp}

In the experiments, we employed four public datasets - PubFig \cite{Kumar2009}, Cifar-100 \cite{Krizhevsky2009}, Caltech-256 \cite{Griffin2007} and Animals with Attributes (AwA) \cite{Lampert2009}. These datasets are designed to pose different visual challenges in terms of illumination effects, scales, and viewpoints as well as support more than 120,000 objects. \\

\noindent \textbf{Implementation details:} In order to evaluate $p(c'_m|d_\text{test})$, 1-vs-all classification is performed. Unless specified, the PubFig, Cifar-100 and Caltech-256 dataset features are extracted using the Pyramid Histogram of Gradient (PHOG) with $3$ pyramid levels, 180$^\circ$ angle and $20$ bins.  Specifically, we use the PHOG from \cite{Bosch2007,bosch2007representing}. However, we did not concatenate all the PHOG descriptors found. Instead, we put all these features in a codebook learning mechanism using the RF algorithm \cite{Moosmann_Nowak_Jurie_2008, WLHoo}. Therefore, we can obtain a set of HOG descriptors that quantize shape information locally and globally, by the nature of the PHOG. The RF codebook can learn image shapes as a whole, as well as the local patch characteristic. For the RF codebook, it is learned using $10$ trees and $100$ leafnodes.

\subsection{PubFig}

The PubFig or Public Figures Face Database has a total of 58797 images of 200 celebrities faces. We used identical subsets as in \cite{Parikh2011} where $8$ random identities are extracted with each class of $100$ images. The pLSA model is built using $K = 11$, similar to the number of attributes in \cite{Parikh2011}. In addition to the PHOG features, we also re-implement our framework using features identical to \cite{Parikh2011}, which is a combination of GIST features and color histograms. We employ the class relationship as in \cite{Parikh2011} to find the $\mathbb{T}_u$.  However, the optimum nearest seen classes pair between the unseen classes are chosen, and we assume the ($\succ$) relationship in \cite{Parikh2011} is similar to our ($\sim$) relationship.

Table \ref{tab:PubFig1} shows that our proposed method has better accuracy (PHOG: $67.83\%$; GIST + color histogram: $69.52\%$), compared to Lampert \etal~~\cite{Lampert2009} that uses the binary attributes, and Parikh and Grauman~\cite{Parikh2011} that uses the relative attributes. Our results are achieved without the annotation required in \cite{Lampert2009,Parikh2011}. When the number of unseen classes $q$ is increased, there is a consistent drop in the system accuracy from $67.83\%$ to $51.30\%$ for PHOG features, and from $69.52\%$ to $57.37\%$ for GIST + color histogram features. This is expected as when the number of unseen classes increases, the system accuracy decreases due to the tradeoffs between computational complexity and system accuracy. 

\begin{table*}[!t]
\caption{PubFig dataset: Performance evaluation (\%) of the proposed method in different numbers of unseen class, $q$ and comparison to state-of-the-art methods.}
\label{tab:PubFig1}
\centering
\renewcommand{\arraystretch}{1.9}
\resizebox{0.9\linewidth}{!}{
\begin{tabular}{|c||c||c||c||c||c||c||c||c|}
\hline
\multirow{3}{*}{Features} & \multicolumn{6}{|c||}{Our
Proposed Method} &\multirow{3}{*}{Binary Attributes} &
\multirow{3}{*}{Relative attributes} \\
\cline{2-7} 
\cline{2-7}
& \multicolumn{6}{|c||}{Number of Unseen Class, $q$} & \multirow{3}{*}{\cite{Lampert2009}} & \multirow{3}{*}{\cite{Parikh2011}}\\
\cline{2-7}
& \textbf{0} & 1 & 2 & 3 & 4 & 5 & & \\
\hline
\hline
PHOG & \textbf{67.83} & 58.89 & 54.99 & 54.35 & 51.65 & 51.30 & N/A & N/A \\
\hline
GIST+color histogram & \textbf{69.52} & 67.09 & 64.25 & 62.55 & 59.52 & 57.37 & 37.00 & 62.00 \\
\hline
\end{tabular}}
\end{table*}

\begin{figure}[!t]
\centering
\includegraphics[height=0.65\linewidth, width=1\linewidth]{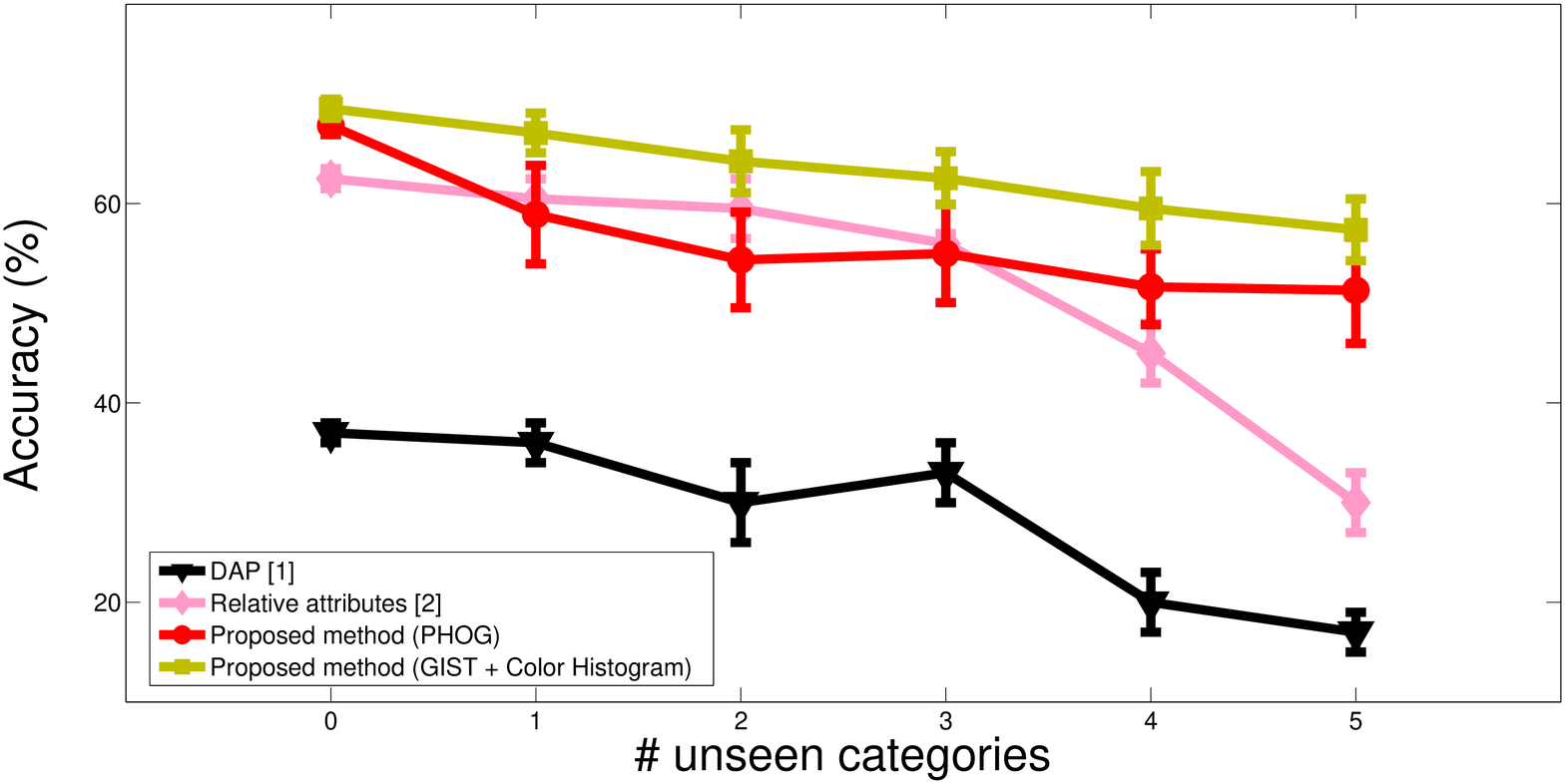}
\caption{Consistency Test: Comparison of the proposed method and the state-of-the-art solutions \cite{Parikh2011, Lampert2009} in the PubFig dataset with different number of unseen categories, $q$. }
\label{fig:PubFig2}
\end{figure}

We performed a consistency test where we tested the accuracy of our proposed method and \cite{Parikh2011,Lampert2009} across different $q$. Figure \ref{fig:PubFig2} shows that the proposed method has a better consistency (PHOG: $\pm$7$\%$; GIST + color histogram: $\pm$10$\%$) in comparison with \cite{Lampert2009}($\pm$17$\%$) and \cite{Parikh2011}($\pm$23$\%$). Also, \cite{Lampert2009} performed the worst in terms of accuracy while \cite{Parikh2011} performed the worst in terms of consistency. Such results have shown the effectiveness and consistency of our proposed algorithm to handle the intra-class variation problem as opposed to the extensive attributes annotation in \cite{Parikh2011,Lampert2009}.

\subsection{Cifar-100}

\begin{table} [!t] 
\caption{Cifar-100: Comparison of the proposed method and the state-of-the-art methods in terms of accuracy (\%).}
\label{tab:Cifar-100}
\centering
\renewcommand{\arraystretch}{1.9}
\resizebox{1\linewidth}{!}{
\begin{tabular}{|c||c||c||c||c||c||c|}
\hline
\multicolumn{5}{|c||}{Our Proposed Method} & \multirow{4}{*}{Sparse Coding} &
\multirow{4}{*}{Beyond} \\
\cline{1-5}
\multicolumn{5}{|c||}{Number of Unseen Class, $q$} & \multirow{4}{*}{ \cite{goodfellow2012large}} & \multirow{4}{*}{Spatial Pyramid}\\
\cline{1-5}
\multicolumn{3}{|c||}{0} & & & & \multirow{4}{*}{\cite{jia2012beyond}}\\
\cline{1-3} 
without & \multicolumn{2}{|c||}{\textbf{with HiC concept}} & 1 & 2 & & \\
\cline{2-3}
HiC concept & J-CoFi & \textbf{CoFi} & & & & \\
\hline
\hline
58.13 & 57.79 & \textbf{58.21} & 56.84 & 54.85 & 53.70 & 54.80\\
\hline
\end{tabular}}
\end{table}

The Cifar-100 \cite{Krizhevsky2009} dataset has 100 classes and each class contains 600 images with $32 \times 32$ resolutions. The 100 classes are further grouped into 20 Coarse Class.  Each $\mathbb{C}$ has 5 $\mathbb{F}$, where $q$ of them is(are) unseen. Thus we have a total $s = q \times 20$.  We picked $30$ training images randomly, and the rest are used for testing. In this dataset we use $K = 10$, as $10$ major semantic topics exist in the $\mathbb{C}$, i.e. mammals, size, trees, vehicles, food, household, insects, reptiles, people, and flowers. The dataset is challenging due to its limited resolution and so we only use $2$ pyramid levels for PHOG features, and $50$ codewords per tree in codebook learning. 

Table \ref{tab:Cifar-100} shows that our proposed method with or without the HiC concept performed much better as compared to \cite{goodfellow2012large,jia2012beyond}. Our approach also outperformed \cite{goodfellow2012large,jia2012beyond} when $q = 2$. When $q = 2$, there is a total of 40 unseen $\mathbb{F}$ when training the classifier. However, our approach was still able to achieve $54.85\%$ accuracy in comparison to $53.70\%$\cite{goodfellow2012large} and $54.80\%$ \cite{jia2012beyond} where in both approaches, $q = 0$ (no unseen classes). In addition, the computational cost of our proposed method is lower, as we only employed a small number of training images. 

Similar to the PubFig dataset, we also observed that when using fewer seen classes in the learning process, the accuracy drops. But, the accuracy differences between $q = 1$ and $q = 2$ only differ by a fraction of $\pm2\%$ even when the difference number of $u$ is large (the total unseen class here is $q \times 20$). 
This indicates that our proposed method is robust as it is capable to handle the Cifar-100 dataset with very tiny ($30 \times 30$) images that causes the collected features vector to be very similar. Besides, in comparison with the three different codebook learning strategies, the CoFi codebook method performs the best as it utilized both $\mathbb{C}_i$ and $\mathbb{F}_i$, simultaneously in the RF tree node splitting.

\begin{table}[h]
\caption{Caltech-256 dataset: Performance evaluation (\%) of the proposed method in different numbers of unseen class, $q$.}
\label{tab:caltech256}
\centering
\renewcommand{\arraystretch}{1.5}
\resizebox{1\linewidth}{!}{
\begin{tabular}{|c||c||c||c||c||c||c||c|}
\hline
\multicolumn{8}{|c|}{Number of Unseen Class, $q$ } \\
\hline
\hline
\multicolumn{3}{|c||}{0} & & & & & \\
\cline{1-3}
{\bf without} & \multicolumn{2}{|c||}{with HiC concept} &1 & 2 & 3 & 4 & 5 \\
\cline{2-3}
{\bf HiC concept} & J-CoFi & CoFi & & & & & \\
\hline
\hline
\textbf{67.72} & 64.60 & 65.65 & 52.14 & 51.49 & 51.86 & 52.13 & 51.32\\
\hline
\end{tabular}}
\end{table}

\begin{table*}[!t]
\caption{Coarse Class, $\mathbb{C}$ for selected Caltech-256 dataset.}
\label{table:new_coarseclass_Caltech256}
\centering
\renewcommand{\arraystretch}{1.2}
\resizebox{1\linewidth}{!}{
\begin{tabular}{|p{3cm}||p{9cm}|}
\hline
Coarse & \multicolumn{1}{c|}{Caltech-256 class} \\
Class, $\mathbb{C}$ & \multicolumn{1}{c|}{(Fine Class, $\mathbb{F}$)} \\
\hline
\hline
household electrical devices & binoculars, boom-box, bread maker, calculator, cd, computer keyboard, computer monitor, computer mouse, floppy-disk, head-phones, iPod, joystick, laptop, light bulb, megaphone, microwave, palm-pilot, paper-shredder, PCI-card, photocopier, refrigerator, rotary-phone, toasters, treadmill, tripod, VCR, video-projector, washing machine\\
\hline
household furniture & bathtub, chandelier, chess-board, desk-globe, doorknob, ewer, flashlight, hammock, hot-tub, hourglass, mailbox, mattress, menorah, picnic table \\
\hline
large man-made outdoor things & Buddha, Eiffel-tower, golden-gate-bridge, light-house, minaret, pyramid, skyscraper, smokestack, teepee, tower-Pisa, windmill\\
\hline
medium mammals & dog, duck, elk, goat, goose, llama, minotaur, penguin, porcupine, raccoon, skunk, swan, unicorn, zebra, greyhound \\
\hline
vehicles & blimp, bulldozer, cannon, canoe, car-tire, covered-wagon, fighting-jet, fire-truck, helicopter, hot-air-ballon, kayak, ketch, license-plate, motorbikes, mountain-bike, pram, school-bus, segway, self-propelled-lawn-mower, snowmobile, speedboat, steering-wheel, touring-bike, tricycles, wheelbarrow, airplanes, car-side \\
\hline
household daily items & beer-mug, chopsticks, coffee-mug, knife, spoon, stained-glass, paperclip, paper-shredder, coins, dice, drinking-straw, dumb-bell, fire-extinguisher, frying-pan, ladder, pez-dispenser, playing-card, roulette-wheel, screwdriver, Swiss-army-knife, tweezer, umbrella \\
\hline
sports & baseball-bat, baseball-glove, baseball-hoop, billiards, bowling-ball, bowling-pin, boxing-glove, football-helmet, Frisbee, golf-ball, skateboard, soccer-ball, tennis-ball, tennis-court, tennis-racket, yo-yo \\
\hline
wears & cowboy-hat, diamond-ring, eyeglasses, football-helmet, necktie, sneaker, socks, top-hat, t-shirt, human-wear, wielding-mask, yarmulke, tennis-shoes, saddle, stirrups\\
\hline
musical instruments & electric-guitar, French-horn, grand-piano, guitar-pick, harmonica, harp, harpsichord, mandolin, sheet-music, tambourine, tuning-fork, xylophone\\ 
\hline
\end{tabular}}
\end{table*}

\begin{figure*}[t]
\centering
\subfigure[Proposed Method]{\includegraphics[height=0.31\linewidth, width=0.31\linewidth]{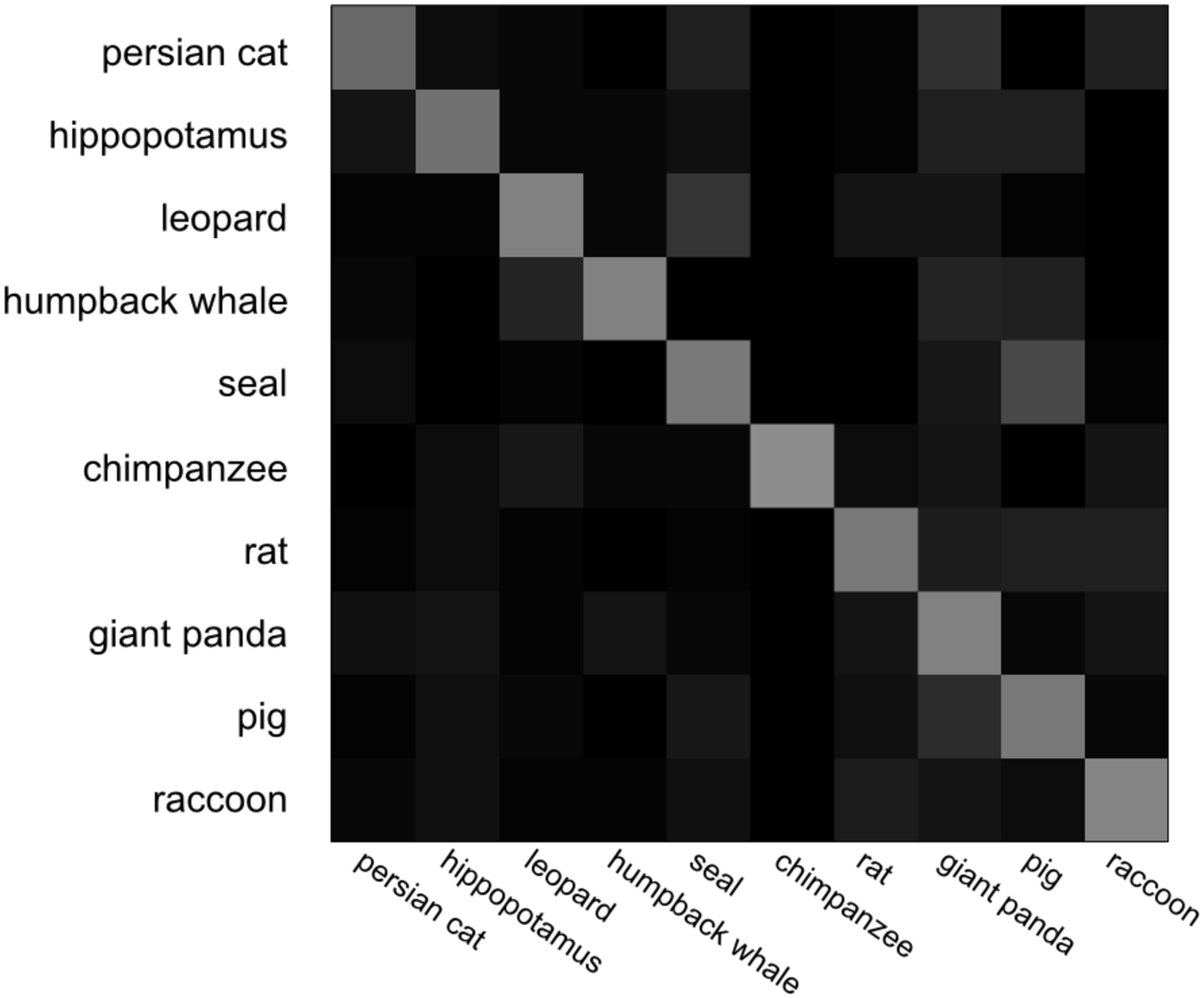}}
\subfigure[IAP \cite{Lampert2014}]{\includegraphics[height=0.31\linewidth, width=0.31\linewidth]{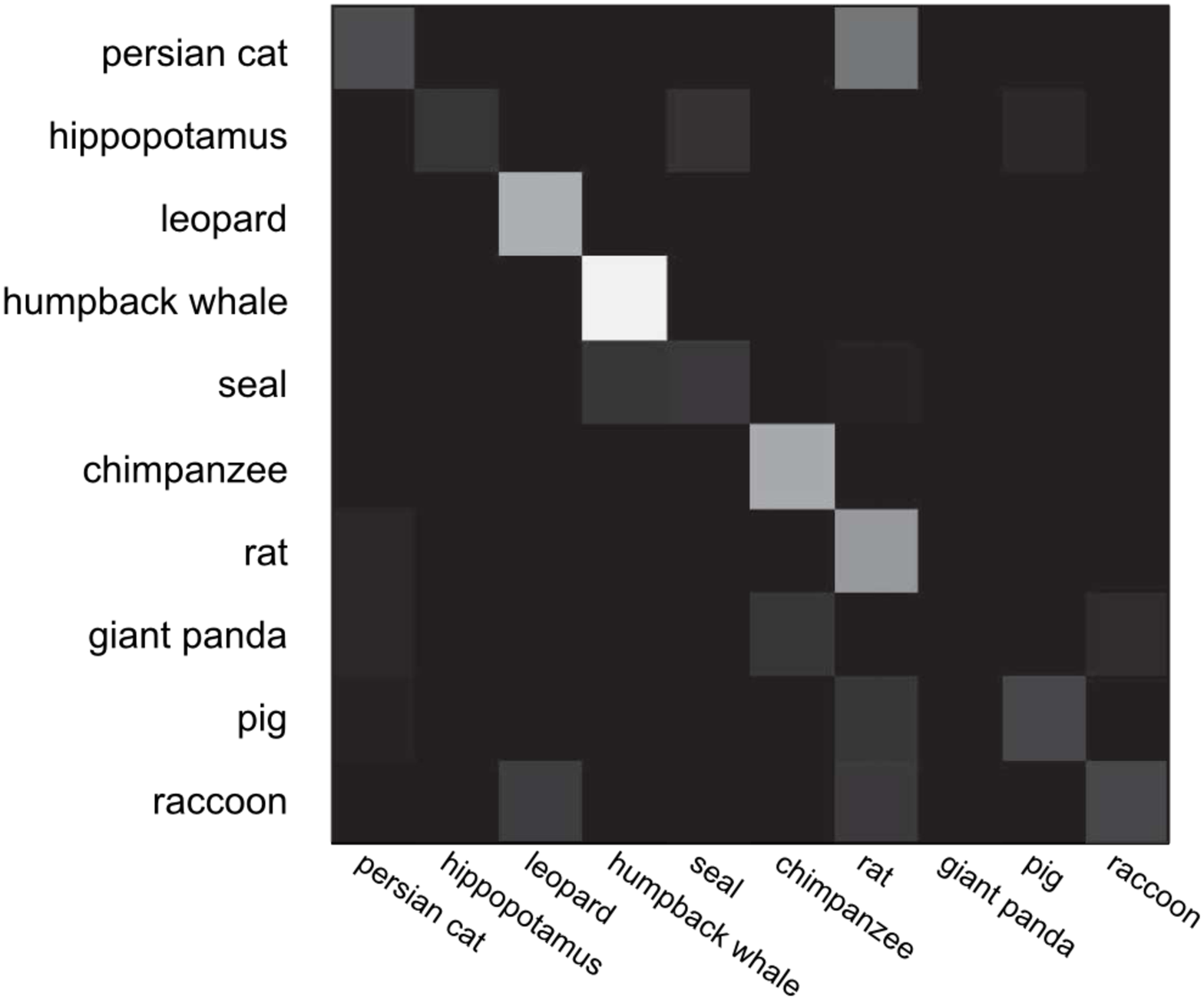}}
\subfigure[DAP \cite{Lampert2014}]{\includegraphics[height=0.31\linewidth, width=0.31\linewidth]{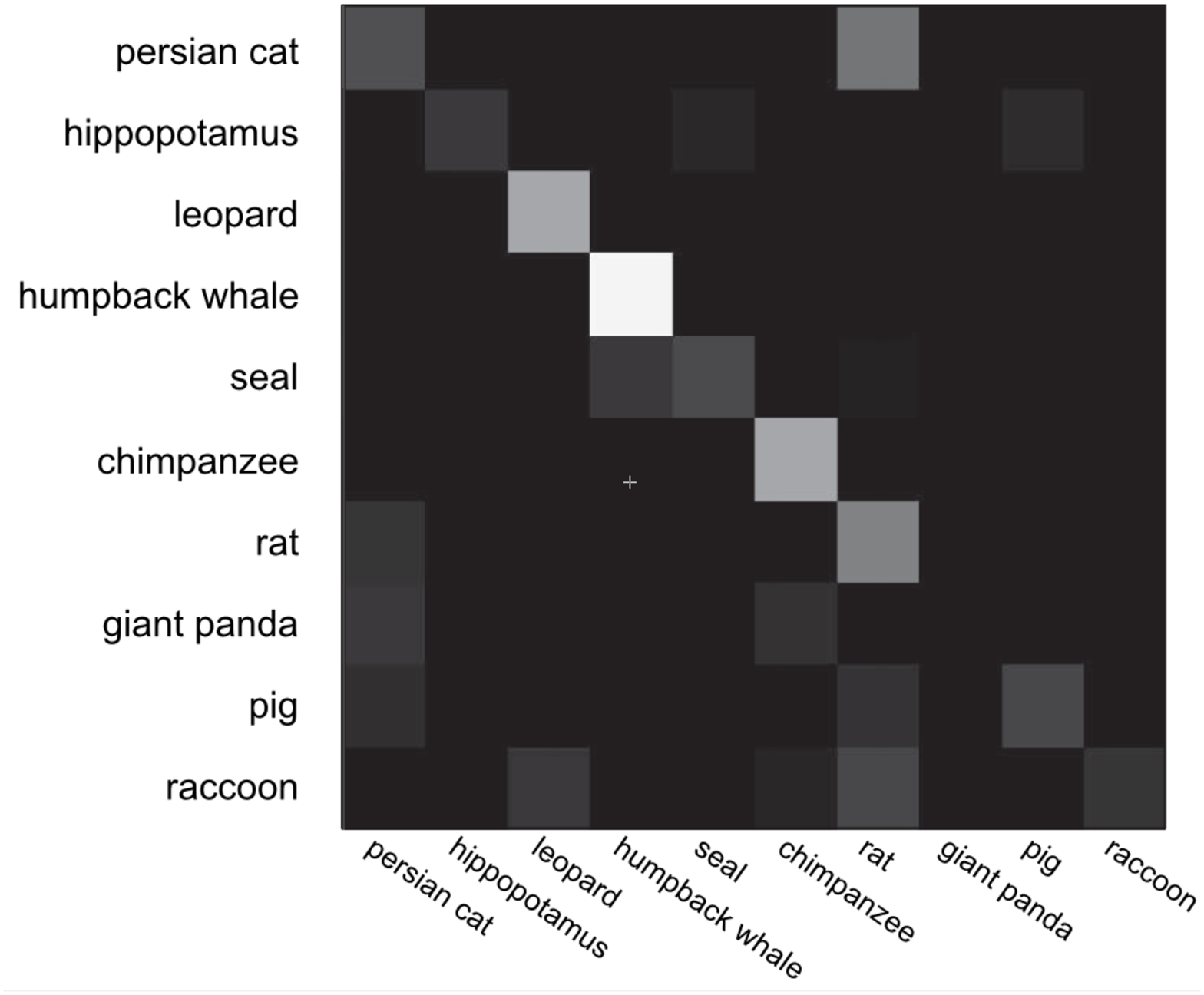}}
\subfigure{\includegraphics[height=0.33\linewidth, width=0.03\linewidth]{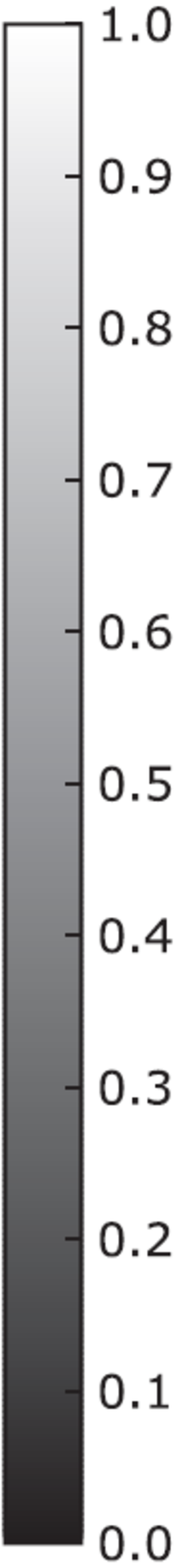}}
\caption{AwA Dataset: Confusion matrix between our proposed method, IAP \cite{Lampert2014} and DAP \cite{Lampert2014}.}
\label{fig:AwA_CM}
\end{figure*}

\subsection{Caltech-256}

The Caltech-256 dataset \cite{Griffin2007} consists of $30607$ images grouped into $256$ object classes and a background class. Unfortunately, it does not provide any $\mathbb{C}$ concepts in the dataset. Therefore, we group the classes manually to $\mathbb{C}$ similar to Cifar-100, except for some specific classes where we introduce new $\mathbb{C}$. In Table \ref{table:new_coarseclass_Caltech256}, we show the distribution of the selected Caltech-256 classes with 5 existing $\mathbb{C}$ as in Cifar-100 and 4 newly introduced $\mathbb{C}$. Only 158 of the total Caltech-256 classes are grouped because some object categories belong to a $\mathbb{C}$ that had very few $\mathbb{F}$ members. For this dataset, the total $u$ are $q \times 9$. 

Table \ref{tab:caltech256} shows minor fluctuations compared to the PubFig and Cifar-100 results when different $q$ values are employed. For classification settings that have $q = 0$, interestingly, the proposed method performs better without applying the HiC concept. We found that this may be due to 1) $\mathbb{F}$ in some $\mathbb{C}$ are semantically related but have low visual similarity (\ie~in this context, the visual similarity is referring to the visual appearance of the object class), \eg~\textquoteleft computer keyboard', \textquoteleft computer monitor' and \textquoteleft computer mouse', which belong to the $\mathbb{C}$ = \textquoteleft household electrical devices'; 2) introducing $\mathbb{C}$ tree to the codebook did not help in boosting the codebook discriminating power, which might be due to the low visual similarity among $\mathbb{F}$ in some $\mathbb{C}$ as well; and 3) the complexity of the objects in Caltech-256. However, the zero-shot learning still provides reasonable results. For this dataset we did not perform any comparisons as there are only 158 classes extracted. 

\subsection{Animal with Attributes (AwA)} 

AwA is an object dataset of animal classes with corresponding attributes attached to each class. There are a total of 50 animal classes and 85 attributes in the dataset. We use similar experimental settings as in \cite{Lampert2009,Lampert2014}, where same features and partitions of seen (\ie~ 40) and unseen (\ie~10) classes were employed. To build the $\mathbb{C}$ and $\mathbb{F}$ relationships, we adopt the attributes relationships and pick the attributes that have $q \leq 2$, and have the lowest number of $\mathbb{F}$ possible. As a result, we grouped 8 $\mathbb{C}$ and each $\mathbb{C}$ has 6 to 12 $\mathbb{F}$, as shown in  Table \ref{table:new_coarseclass_AwA}.

\begin{table*}[!t]
\caption{Coarse Class, $\mathbb{C}$ for AwA dataset. The object classes in {\bf bold} are the predefined unseen classes in \cite{Lampert2009}.}
\label{table:new_coarseclass_AwA}
\centering
\renewcommand{\arraystretch}{1.5}
\resizebox{1\linewidth}{!}{
\begin{tabular}{|p{1.5cm}|p{8cm}|}
\hline
Coarse & \multicolumn{1}{c|}{AwA class} \\
Class, $\mathbb{C}$ & \multicolumn{1}{c|}{(Fine Class, $\mathbb{F}$)} \\
\hline
\hline
hooves  & antelope, horse, moose, ox, sheep, rhinoceros, giraffe, buffalo, zebra, deer, {\bf pig}, cow \\
\hline
weak    & Siamese cat, {\bf Persian cat}, skunk, mole, sheep, hamster, rabbit, bat, chihuahua, mouse \\
\hline
grazer  & antelope, horse, moose, spider monkey, elephant, ox, sheep, hamster, rhinoceros, rabbit, giraffe, buffalo, zebra, {\bf giant panda}, deer, mouse, cow \\
\hline
stalker & grizzly bear, German shepherd, Siamese cat, tiger, {\bf leopard}, fox, wolf, bobcat, lion, polar bear \\
\hline
flippers & killer whale, blue whale, {\bf humpback whale}, {\bf seal}, otter, walrus, dolphin \\
\hline
strainteeth & killer whale, beaver, blue whale, {\bf hippopotamus}, {\bf humpback whale}, walrus \\
\hline
hibernate & grizzly bear, beaver, skunk, mole, fox, hamster, squirrel, bat, {\bf rat}, bobcat, mouse, polar bear, {\bf raccoon} \\
\hline
bipedal & grizzly bear, spider monkey, gorilla, {\bf chimpanzee}, squirrel, bat, {\bf giant panda}, polar bear\\
\hline
\end{tabular}}
\end{table*}

Herein, our proposed method achieved an accuracy of $49.65\%$. This is better as compared to the DAP and IAP \cite{Lampert2014}, which achieved $41.4\%$ and $42.2\%$ respectively; to M2LATM \cite{Fu2014Learning} that obtain $41.3\%$; and attribute/hierarchical label embedding (AHLE) \cite{akata2013label} that achieved $43.5\%$. We also show the confusion matrix of the 10 test classes in Figure \ref{fig:AwA_CM}. We observe that our proposed method has better average classification results compared to DAP and IAP \cite{Lampert2014}. Though our proposed method does not predict the `humpback whale' class as well as DAP and IAP, but we achieve better accuracy in the `giant panda' class, which leads to better overall accuracy. These results benefit from the HiC concept defined for AwA, where `giant panda' class is the only $u$ in $\mathbb{C} = \{\text{`grazer'}\}$. Note that the `humpback whale' class resides in $\mathbb{C} = \{\text{`flippers',`strainteeth'}\}$, and both $\mathbb{C}$ contains more than one $u$. Therefore we observe the accuracy drops in the `humpback whale' class. The same situation applies to the `rat' class and `raccoon' class as they share the same $\mathbb{C} = \{\text{`hibernate'\}}$.

\section{Discussion}
\label{sec:discussion}

In this paper, we compared our proposed method with 4 public datasets and achieves better performance compared to state-of-the-art methods for zero-shot learning.  Even in the conventional classification problem where training images for all object classes are available, we still manage to get state-of-the-art accuracy in PubFig and Cifar-100 datasets.

In the conducted experiments, there are some cases where the predicted $\mathbb{T}_u$ is redundant. That is, if a lower number of $K$ is chosen, the numbers of possible $M$ will be reduced as well, and hence there is a possibility to obtain similar $\mathbb{T}_u$ for different $u$, which is a redundant representation.  In order to handle this issue, we employ a large number of $K$ in the experiments to minimize the probability of $\mathbb{T}_c$ to be redundant.

Based on the experiments in Caltech-256 dataset, we are aware that the classification accuracy is fluctuating due to the quality of $\mathbb{F}$ collection under each $\mathbb{C}$.  Though, the $\mathbb{F}$ within $\mathbb{C}$ is grouped based on the semantic relationship; these $\mathbb{F}$ might be visually dissimilar.  This limitation is likely to be solved by introducing a middle-level class group to extend the $\mathbb{F}$ within the $\mathbb{C}$ to some high-visual similarity group, \eg~we can group and extend $\mathbb{F}$: `head-phones', `rotary-phones' and `megaphone' in $\mathbb{C}$: `household electrical devices' to $\mathbb{C}$: `phones'. When we pick the random $\alpha_u$ to model $u =\{\text{`megaphone'}\}$, the `head-phones' and `rotary-phones' will have priority as the related $s$. Nonetheless, our future work includes introducing tighter relationship between the Fine Class in same the Coarse Class so that better performance can be achieved.

\bibliographystyle{IEEEbib}
\bibliography{refs}

\begin{thebibliography}{10}

\bibitem{Lampert2009}
C.H. Lampert, H.~Nickisch, and S.~Harmeling,
\newblock ``Learning to detect unseen object classes by between-class attribute
  transfer,''
\newblock in {\em IEEE Conference on Computer Vision and Pattern Recognition},
  2009, pp. 951--958.

\bibitem{Parikh2011}
D.~Parikh and K.~Grauman,
\newblock ``Relative attributes,''
\newblock in {\em IEEE International Conference on Computer Vision}, 2011, pp.
  503 --510.

\bibitem{Kumar2009}
N.~Kumar, A.C. Berg, P.N. Belhumeur, and S.K. Nayar,
\newblock ``Attribute and simile classifiers for face verification,''
\newblock in {\em IEEE International Conference on Computer Vision}, 2009, pp.
  365--372.

\bibitem{rohrbach2011evaluating}
M.~Rohrbach, M.~Stark, and B.~Schiele,
\newblock ``Evaluating knowledge transfer and zero-shot learning in a
  large-scale setting,''
\newblock in {\em IEEE Conference on Computer Vision and Pattern Recognition},
  2011, pp. 1641--1648.

\bibitem{frome2013devise}
A.~Frome, G.~S Corrado, J.~Shlens, S.~Bengio, J.~Dean, T.~Mikolov, M.A.
  Ranzato, and T.~Mikolov,
\newblock ``Devise: A deep visual-semantic embedding model,''
\newblock in {\em Advances in Neural Information Processing Systems}, 2013, pp.
  2121--2129.

\bibitem{mensink2012metric}
T.~Mensink, J.~Verbeek, F.~Perronnin, and G.~Csurka,
\newblock ``Metric learning for large scale image classification: Generalizing
  to new classes at near-zero cost,''
\newblock in {\em European Conference on Computer Vision}, pp. 488--501.
  Springer, 2012.

\bibitem{Hoo2013}
W.L. Hoo and C.S. Chan,
\newblock ``Plsa-based zero-shot learning,''
\newblock in {\em 20th IEEE International Conference on Image Processing},
  2013, pp. 4297--4301.

\bibitem{fei2005bayesian}
L.~Fei~Fei and P.~Perona,
\newblock ``A bayesian hierarchical model for learning natural scene
  categories,''
\newblock in {\em IEEE Conference on Computer Vision and Pattern Recognition},
  2005, vol.~2, pp. 524--531.

\bibitem{Griffin2007}
G.~Griffin, A.~Holub, and P.~Perona,
\newblock ``Caltech-256 object category dataset,''
\newblock {\em Caltech Technical Report, No. CNS-TR-2007-001.}, 2007.

\bibitem{Bosch2007}
A.~Bosch, A.~Zisserman, and X.~Muoz,
\newblock ``Image classification using random forests and ferns,''
\newblock in {\em IEEE International Conference on Computer Vision}, 2007, pp.
  1--8.

\bibitem{Moosmann_Nowak_Jurie_2008}
F.~Moosmann, E.~Nowak, and F.~Jurie,
\newblock ``Randomized clustering forests for image classification.,''
\newblock {\em IEEE Transactions on Pattern Analysis and Machine Intelligence},
  vol. 30, pp. 1632 --1646, 2008.

\bibitem{ferrari2007learning}
V.~Ferrari and A.~Zisserman,
\newblock ``Learning visual attributes,''
\newblock in {\em Advances in Neural Information Processing Systems}, 2007, pp.
  433--440.

\bibitem{Lampert2014}
C.H. Lampert, H.~Nickisch, and S.~Harmeling,
\newblock ``Attribute-based classification for zero-shot visual object
  categorization,''
\newblock {\em IEEE Transactions on Pattern Analysis and Machine Intelligence},
  vol. 36, no. 3, pp. 453--465, 2014.

\bibitem{fu2012attribute}
Y.~Fu, T.M. Hospedales, T.~Xiang, and S.~Gong,
\newblock ``Attribute learning for understanding unstructured social
  activity,''
\newblock in {\em European Conference on Computer Vision}, pp. 530--543.
  Springer, 2012.

\bibitem{Fu2014Learning}
Y.~Fu, T.M. Hospedales, T.~Xiang, and S.~Gong,
\newblock ``Learning multimodal latent attributes,''
\newblock {\em IEEE Transactions on Pattern Analysis and Machine Intelligence},
  vol. 36, no. 2, pp. 303--316, 2014.

\bibitem{guillaumin2010multimodal}
M.~Guillaumin, J.~Verbeek, and C.~Schmid,
\newblock ``Multimodal semi-supervised learning for image classification,''
\newblock in {\em IEEE Conference on Computer Vision and Pattern Recognition},
  2010, pp. 902--909.

\bibitem{Palatucci2009}
M.~Palatucci, D.~Pomerleau, G.~Hinton, and T.~Mitchell,
\newblock ``Zero-shot learning with semantic output codes,''
\newblock in {\em Advances in Neural Information Processing Systems}, 2009, pp.
  1410--1418.

\bibitem{silberer2013models}
C.~Silberer, V.~Ferrari, and M.~Lapata,
\newblock ``Models of semantic representation with visual attributes,''
\newblock in {\em Proceedings of the 51th Annual Meeting of the Association for
  Computational Linguistics}, 2013, pp. 572--582.

\bibitem{WLHoo}
W.L. Hoo, T-K Kim, Y.~Pei, and C.S. Chan,
\newblock ``Enhanced random forest with image/patch-level learning for image
  understanding,''
\newblock in {\em Proceedings of the 22nd International Conference on Pattern
  Recognition}, 2014, pp. 3434--3439.

\bibitem{Sivic_Russell_Efros_Zisserman_Freeman_2005}
J.~Sivic, B.~Russell, A~Efros, A.~Zisserman, and W.~Freeman,
\newblock ``Discovering objects and their location in images,''
\newblock in {\em IEEE Conference on Computer Vision and Pattern Recognition},
  2005, pp. 370--377.

\bibitem{Krizhevsky2009}
A.~Krizhevsky,
\newblock ``Learning multiple layers of features from tiny images,''
\newblock Tech. {R}ep., 2009.

\bibitem{bosch2007representing}
A.~Bosch, A.~Zisserman, and X.~Munoz,
\newblock ``Representing shape with a spatial pyramid kernel,''
\newblock in {\em ACM International Conference on Image and Video Retrieval},
  2007, pp. 401--408.

\bibitem{goodfellow2012large}
I.~Goodfellow, A.~Courville, and Y.~Bengio,
\newblock ``Large-scale feature learning with spike-and-slab sparse coding,''
\newblock in {\em International Conference on Machine Learning}, 2012, pp.
  1439--1446.

\bibitem{jia2012beyond}
Y.~Jia, C.~Huang, and T.~Darrell,
\newblock ``Beyond spatial pyramids: Receptive field learning for pooled image
  features,''
\newblock in {\em IEEE Conference on Computer Vision and Pattern Recognition},
  2012, pp. 3370--3377.

\bibitem{akata2013label}
Z.~Akata, F.~Perronnin, Z.~Harchaoui, and C.~Schmid,
\newblock ``Label-embedding for attribute-based classification,''
\newblock in {\em IEEE Conference on Computer Vision and Pattern Recognition},
  2013, pp. 819--826.

\end{thebibliography}

\vfill

\end{document}